\documentclass[10pt,twocolumn,letterpaper]{article}

\usepackage{iccv}
\usepackage{times}
\usepackage{epsfig}
\usepackage{graphicx}
\usepackage{amsmath}
\usepackage{amssymb}
\usepackage{balance}
\usepackage{mathtools}
\usepackage{booktabs}
\usepackage{sidecap}
\usepackage{soul}
\usepackage{algorithm}
\usepackage[noend]{algorithmic}
\usepackage[table]{xcolor}
\usepackage{enumitem}
\setlist{leftmargin=5mm}
\newtheorem{example}{Example}

\newtheorem{definition}{Definition}
\newtheorem{proposition}{Proposition}
\newtheorem{lemma}{Lemma}

\newcommand{\improvement}[1]{\textcolor[rgb]{0.22,0.463,0.114}{\footnotesize{\textbf{#1}}}}
\newcommand{\reduction}[1]{\textcolor[rgb]{0.763,0.111,0.114}{\footnotesize{\textbf{#1}}}}

\usepackage[pagebackref=true,breaklinks=true,letterpaper=true,colorlinks,bookmarks=false]{hyperref}

 \iccvfinalcopy 

\def\iccvPaperID{3730} 

\ificcvfinal\fi

\begin{document}

\title{Towards credible visual model interpretation with path attribution}

\author{Naveed Akhtar\\
\and
Muhammad A. A. K. Jalwana\\
}

\maketitle
\ificcvfinal\thispagestyle{empty}\fi

\begin{abstract}
Originally inspired by game-theory, path attribution framework stands out among the post-hoc model interpretation tools due to its axiomatic nature. However, recent developments show that this framework can still suffer from  counter-intuitive results. 
Moreover, specifically for  deep visual models, the existing path-based methods also fall short on  conforming to the original intuitions that are the basis of the claimed axiomatic properties of this framework. We address these problems with a systematic  investigation, and pinpoint the conditions in which the counter-intuitive results can be avoided for deep visual model interpretation with the path attribution strategy. 
We also devise a scheme to preclude the conditions in which visual model interpretation can invalidate the axiomatic properties of path attribution.   
These insights are combined into a  method that enables reliable visual model interpretation. Our findings are establish empirically with multiple datasets, models and evaluation metrics. Extensive experiments show a  consistent performance gain of our method over the baselines. 
\end{abstract}

\vspace{-3mm}
\section{Introduction}
\label{sec:intro}
\vspace{-1mm}
Deep learning is fast approaching the maturity where it can be commonly deployed in safety-critical applications~\cite{Nature1}, \cite{akhtar2021advances}, \cite{rudin2019stop}. However, its black-box nature presents a major concern for its use in high-stake applications~\cite{DeepMind}, \cite{agarwal2021neural}, \cite{blazek2021explainable}, and its ethical use in general \cite{vinuesa2020role}, \cite{vinuesa2021interpretable}. These facts have led to the development of numerous techniques of explaining  deep learning  models~\cite{agarwal2021neural}, \cite{akhtar2023rethinking}, \cite{chen2020concept}, \cite{jalwana2021cameras}, \cite{selvaraju2017grad},  \cite{Simonyan2014Deep}, \cite{Sundararajan2017Axiomatic}. Whereas rendering the  models intrinsically explainable is an active parallel research direction~\cite{agarwal2021neural}, \cite{blazek2021explainable}, \cite{chen2020concept}, \cite{koh2020concept},  \textit{post-hoc} model interpretation methods are currently highly popular. These methods are used to explain the predictions of already deployed models. 


Path attribution methods~\cite{Erion2021Improving}, \cite{kapishnikov2021guided}, \cite{pan2021explaining}, \cite{Sundararajan2017Axiomatic} hold a special place among the post-hoc interpretation techniques due to their clear theoretical foundations. These methods compute attribution scores (or simply the  \textit{attributions}) for the input features to quantify their importance for  model prediction, where the attributions and the methods follow certain desirable axiomatic properties~\cite{Sundararajan2017Axiomatic}. These properties emerge from the cooperative game-theory roots of the path attribution framework~\cite{friedman2004paths}, \cite{Sundararajan2017Axiomatic}. 

\begin{figure}[t]
    \centering
    \includegraphics[width = 0.47\textwidth]{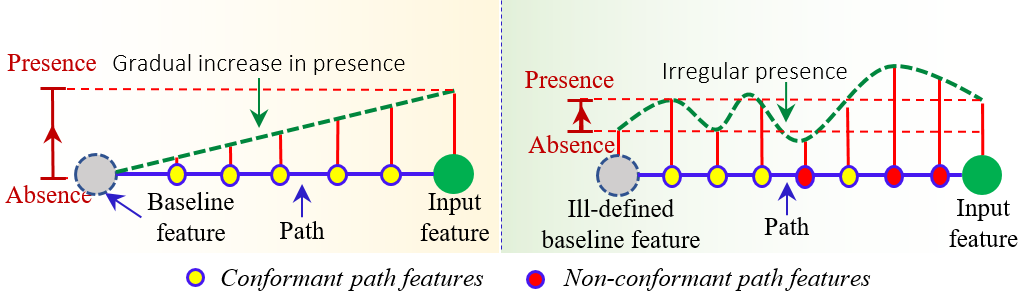}
    \caption{\textbf{Left}: The  cooperative game-theory inspired path attribution framework defines a path between the input feature `absence' and its `presence' using a baseline. 
    \textbf{Right}: Abstract nature of feature absence/presence in visual modelling not only causes  ill-defined baselines, but can also result in non-conformant intermediate  path features, which violate the assumption that  they  reside within the extremities of input feature absence and presence.} 
    \label{fig:teaser}
    \vspace{-3mm}
\end{figure}

To compute the attribution score for an input feature, the path attribution framework defines a \textit{baseline}, and a \textit{path} between the input and the baseline. Following the original game-theory  intuitions~\cite{aumann2015values}, \cite{friedman2004paths}, \cite{Sundararajan2017Axiomatic}, the baseline signifies `absence' of the input feature, and the path gradually flows from this absence to the  `presence' of the feature in the input. This intuition has direct implications for the desirable axiomatic properties of the framework. However, an unambiguous definition of `feature absence' still eludes visual modelling~\cite{Erion2021Improving},  \cite{pan2021explaining}, \cite{Sturmfels2020Visualizing}. This leads to ill-defined baselines, which in turn also results in problematic features on the path between the baseline and the input. These features violate the original assumption  that their existence is between the extremities of the absence and presence of the input feature - see Fig.~\ref{fig:teaser}.      
Besides, Srinivas~\etal~\cite{Srinivas2019Full} also showed that the existing path attribution methods for deep learning can compute counter-intuitive results even when they satisfy their claimed axiomatic properties. 
  
In this work, we address the above-noted shortcomings of the path attribution framework for interpreting deep visual models. The key contributions of this paper are: 

\vspace{-0.5mm}
\begin{itemize}[noitemsep,topsep=0pt]
  \item With a systematic theoretical formulation, it pinpoints the conflicts that result in the problems of (\textit{i}) counter-intuitive attributions, (\textit{ii}) ambiguity in the baseline and (\textit{iii})~non-conformant path features; which  collectively compromise the reliability of the path attribution framework for deep visual model interpretation. 
  \item For each of these problems, it proposes a theory-driven solution, which conforms to the original game-theory intuitions of the path attribution framework.   
  \item It combines the solutions into a  well-defined path attribution method to compute reliable attributions.
  \item It thoroughly establishes the efficacy of the proposed method with extensive experiments using multiple models, datasets and evaluation metrics.   
\end{itemize}

\vspace{-3mm}
\section{Related work}
\label{sec:RW}
\vspace{-1.5mm}
Due to the critical need of interpreting deep learning predictions in high-stake applications, techniques to explain deep neural models are gaining considerable research attention. Whereas a stream of works exists that aims at making these models inherently explainable~\cite{chen2019looks}, \cite{brendel2019approximating}, \cite{bohle2021convolutional}, \cite{bohle2022b}, \cite{donnelly2022deformable}, \cite{sarkar2022framework}, \cite{parekh2021framework}, post-hoc  interpretation techniques~\cite{Sundararajan2017Axiomatic}, \cite{slack2021reliable}, \cite{jalwana2021cameras}, \cite{Smilkov2017Smoothgrad} currently dominate the existing related literature. A major advantage of the post-hoc methods is that the interpretation process does not interfere with the model design and training, thereby leaving the model performance unaffected. Our method is also a post-hoc technique, hence we focus on the related literature along this stream.   

Based on the underlying mechanism, we can divide the post-hoc interpretation approaches into three categories. The first is \textit{perturbation-based} techniques~\cite{dabkowski2017real}, \cite{fong2017interpretable}, \cite{ribeiro2016should}, \cite{petsiuk2018rise}, \cite{zeiler2014visualizing}. The central idea of these methods is to interpret model prediction by perturbing the input features and analyzing its effects on the  output. For instance, \cite{petsiuk2018rise}, \cite{ribeiro2016should}, \cite{zeiler2014visualizing} occlude parts of the image to cause the perturbation, whereas \cite{dabkowski2017real}, \cite{fong2017interpretable} optimize for a perturbation mask, keeping in sight the confidence score of the predictions. These methods are particularly relevant to black-box scenarios, where the model details are unavailable. However, since white-box setups are equally practical for the interpretation task, other works also use model information to devise more efficient methods. We discuss these approaches next.  

Among the white-box methods, \textit{activation-based} techniques \cite{selvaraju2017grad}, \cite{jalwana2021cameras}, \cite{chattopadhay2018grad}, \cite{ramaswamy2020ablation}, \cite{jiang2021layercam}, \cite{wang2020score} form the second broad category. These methods commonly interpret the model predictions by weighting the activations of the deeper layers of the network with  model gradients, thereby computing a saliency map for the input features. Though efficient, the resolution mismatch between the deeper layer features of the model and the input image compromises the quality of their results~\cite{jalwana2021cameras}. The third category is that of \textit{backpropagation-based} techniques~\cite{simonyan2013deep}, \cite{shrikumar2017learning}, \cite{Srinivas2019Full}, \cite{Sundararajan2017Axiomatic}, \cite{zhang2018top}, which side steps this issue by fully backpropagating the model gradients to the input for feature  interpretation.

Within the backpropagation-based techniques, a sub-branch of approaches is known as \textit{path attribution methods} \cite{Sundararajan2017Axiomatic}, \cite{Erion2021Improving}, \cite{Sturmfels2020Visualizing}, \cite{pan2021explaining}, \cite{Smilkov2017Smoothgrad}, \cite{kapishnikov2021guided}. These methods are particularly attractive because they exhibit certain desirable axiomatic properties~\cite{Sundararajan2017Axiomatic}, \cite{lundstrom2022rigorous}. Originated in cooperative game-theory~\cite{friedman2004paths}, the central idea of these techniques is to accumulate model gradients w.r.t.~the input over a single~\cite{Sundararajan2017Axiomatic} or multiple~\cite{Erion2021Improving}, \cite{lundstrom2022rigorous} paths formed between the input and a  baseline image. The baseline signifies absence of the input features. Recording  model gradients from the absence to the presence of a feature allows a more desirable non-local estimate of the importance attributed by the model to that feature. 

\vspace{-2mm}
\section{Path attribution framework}
\label{sec:PAF}
\vspace{-1mm}
We contribute to the path attribution framework~\cite{Sundararajan2017Axiomatic}, \cite{lundstrom2022rigorous}. 
This section provides a concise and self-contained 
formal  discussion on the concepts that are relevant to the remaining paper. We refer to \cite{Sundararajan2017Axiomatic} and \cite{lundstrom2022rigorous} for the other details of the broader framework.     

Path attribution methods build on the concepts of \textit{baseline attribution} and \textit{path function}. To formalize these concepts, consider two points $\boldsymbol a, \boldsymbol b \in \mathbb R^n$ that  define a hyper-rectangle $[\boldsymbol a, \boldsymbol b]$ as its opposite vertices. For instance, $\boldsymbol a$ and $\boldsymbol b$ can be (vectorized) black and white images, respectively; that form the hyper-rectangle encompassing the pixel values of images in $\mathbb R^n$. A visual classifier $F$ then belongs to a class of single output functions $\mathcal F:[\boldsymbol a, \boldsymbol b] \rightarrow \mathbb R$. 
The notion of baseline attribution can be  defined as
\vspace{-0.7mm}
\begin{definition}[Baseline attribution]\label{def:BAM}
Given  $F\in\mathcal{F(\boldsymbol a, \boldsymbol b)}$, $\boldsymbol x, \boldsymbol x' \in [\boldsymbol a, \boldsymbol b]$, a  baseline attribution method is a function of the form $\mathcal A: [\boldsymbol a, \boldsymbol b] \times [\boldsymbol a, \boldsymbol b] \times \mathcal{F}(\boldsymbol a, \boldsymbol b) \rightarrow \mathbb R^n$.
\end{definition}
\vspace{-0.7mm}

\noindent In Def.~(\ref{def:BAM}), $\boldsymbol x = [x_1, x_2,...,x_n] \in \mathbb R^n$ denotes the  \textit{input} to the classifier $F$. The vector $\boldsymbol x' \in \mathbb R^n$ is the  \textit{baseline}. For a visual model, the objective of $\mathcal A$ is to estimate the contribution of each pixel $x_i$ of  $\boldsymbol x$ to the model output. The idea of baseline attribution is fundamental to all the path attribution methods. The other key concept for this paradigm is  \textit{path function}, which  can be concisely stated as

\vspace{-0.7mm}
\begin{definition}[Path function]\label{def:pf}
A function $\gamma(\boldsymbol x, \boldsymbol x', \alpha) : [\boldsymbol a, \boldsymbol b] \times [\boldsymbol a, \boldsymbol b] \times [0,1] \rightarrow [\boldsymbol a, \boldsymbol b]$ is a path function, if for a given pair $\boldsymbol x, \boldsymbol x'$, $\gamma(\alpha)\coloneqq \gamma(\boldsymbol x, \boldsymbol x', \alpha)$ is a continuous piece-wise smooth curve from $\boldsymbol x'$ to $\boldsymbol x$.
\end{definition}
\vspace{-0.7mm}

In Def.~(\ref{def:pf}), it is assumed that $\frac{\partial F(\gamma(\alpha))}{\partial x_i} $ exists everywhere. All axiomatic path attribution methods  follow this assumption\footnote{It is assumed by the methods that the Lebesgue measure for the set of points where the function is not defined is 0.}. We can unify these methods as further  specifications of the following broad definition.
\vspace{-0.7mm}
\begin{definition}[Path attribution  methods]\label{def:PbM} For a path function $\gamma(\alpha)$, its path  attribution method solves for
\begin{equation}
    \mathcal{A}(\boldsymbol x, \boldsymbol x', \gamma) = \int\limits_{0}^1 \frac{\partial F(\gamma(\alpha))}{\partial x_i}  \times \frac{\partial \gamma_i(\alpha)}{\partial \alpha} d\alpha, 
    \label{eq:PbM}
\end{equation}
\vspace{-0.7mm}
where the subscript `$i$' indicates the $i^{\text{th}}$ entry of the entity. 
\end{definition}
\vspace{-0.7mm}

Among the path attribution methods, Integrated Gradients (IG)~\cite{Sundararajan2017Axiomatic} is considered the canonical method. It  uses a linear path in Eq.~(\ref{eq:PbM}), i.e.,~$\gamma(\alpha) = \boldsymbol x' + \alpha (\boldsymbol x - \boldsymbol x')$, thereby solving the following for baseline attribution 
\begin{equation} 
\mathcal A_i(\boldsymbol x, \boldsymbol x') = (x_i - x'_i) \times \int\limits_{\alpha=0}^1 \frac{\partial F(\boldsymbol x'+\alpha(\boldsymbol x- \boldsymbol x'))}{\partial x_i} d\alpha,
\label{eq:IG}
\end{equation}
where $x_i'$ is the $i^{\text{th}}$ pixel in the baseline image. In Eq.~(\ref{eq:IG}), the subscript `$i$' in $\mathcal{A}_i(.)$ indicates that the attribution is estimated for a single feature. For simplicity, in the text to follow, we often re-purpose $\mathcal A$ to  refer to  an \textit{attribution map}, s.t.~$\mathcal{A} = \{\mathcal{A}_1, \mathcal{A}_2,...,\mathcal{A}_n\}$. Herein, $\mathcal A_i$ denotes the attribution \textit{score} of the feature $x_i$, e.g., solution to Eq.~(\ref{eq:IG}).  

A systematic use of the baseline and path function enables the path attribution methods to demonstrate  a range of   axiomatic properties~\cite{Sundararajan2017Axiomatic}, \cite{friedman2004paths}, \cite{lundstrom2022rigorous}. \textcolor{black}{We discuss these properties in the context of our contribution in the supplementary material.}
Here, we must formally define one of those properties, called \textit{completeness}, as it is critical to understand the remaining discussion in the main paper.
\vspace{-0.5mm}
\begin{definition}[Completeness]\label{def:comp} For $F\in \mathcal F (\boldsymbol a, \boldsymbol b)$ and $\boldsymbol x, \boldsymbol x' \in [\boldsymbol a, \boldsymbol b]$, we have $\sum_{i = 1}^n \mathcal{A}_i (\boldsymbol x, \boldsymbol x') = F(\boldsymbol x) - F(\boldsymbol x')$.
\end{definition}
\vspace{-0.5mm}
Completeness asserts that a non-zero importance is attributed to a feature  only when that feature  contributes to the output.
From Def.~(\ref{def:BAM})-(\ref{def:PbM}), it is apparent that the  path attribution methods 
rely strongly on (\textit{i}) the baseline $\boldsymbol x'$ and (\textit{ii}) the path used to compute the attribution scores. These two aspects will  remain at the center of our discussion below. 

\vspace{-2mm}
\section{Problems with visual path attribution}
\label{sec:Prob}
\vspace{-1mm}
The pioneering path attribution method in the vision domain, i.e., Integrated Gradients (IG)~\cite{Sundararajan2017Axiomatic} took inspiration from  cooperative game-theory~\cite{friedman2004paths}. In fact,  IG corresponds to a cost-sharing method called  Aumann-Shapley~\cite{aumann2015values}. 
However, Lundstrom et al.~\cite{lundstrom2022rigorous}  recently noted that the class of functions $\mathcal F$ - see \S~\ref{sec:PAF}, \textit{cf.}~Def.~(\ref{def:BAM}) - implemented by  deep learning models, e.g., visual classifiers, behave differently than the corresponding functions in the cooperative game-theory setups. Hence, the path attribution framework needs further investigation for visual model interpretation in regards to its claimed  properties. Below we explicate the peculiar  problems faced when this framework is applied to the  deep visual models.  
For clarity, we defer the proposed solutions to these problems to \S~\ref{sec:Fixes}. 

\vspace{0.7mm}
\noindent{\underline{\bf P1: Counter-intuitive attribution scores:}} Srinivas and Fleuret~\cite{Srinivas2019Full} first highlighted  a critical issue of `counter-intuitive' attribution scores computed by IG~\cite{Sundararajan2017Axiomatic}. We provide an accessible  example below to clearly explain the problem. The examples and  discussion herein  are directly applicable to the modern deep visual models as they are  represented well as piece-wise linear functions~\cite{Srinivas2019Full}. 
\vspace{-1mm}
\begin{example}[Counter-intuitive attribution scores]
\label{example1} Define a piece-wise linear function for an input $\boldsymbol x = [x_1, x_2] \in \mathbb{R}^2$.
\begin{equation*} \label{eq:example1}
F(\boldsymbol x) \!\!= \!\!\left\{ \!\!\!\!
\begin{array}{rc}
x_1+4x_2+1, &       \mathcal U_1 = \{\boldsymbol x ~|~ x_1, x_2 \leq 1 \}\\
4x_1+x_2+2,       & \mathcal U_2 = \{\boldsymbol x ~|~ x_1, x_2 > 1 \}\\
0,       & \rm{otherwise.}
\end{array} \right.
\end{equation*}
\vspace{-0.7mm}
Consider two  points $\boldsymbol x^a = [1.5,1.5]$ and $\boldsymbol x^b = [4,4]$ s.t. $\boldsymbol x^a, \boldsymbol x^b \in \mathcal U_2$. For these points, $x_1$ clearly influences the output more strongly than $x_2$ due to its larger weight, i.e.,~4 vs 1. However,  when we apply Eq.~(\ref{eq:PbM}) for attribution computation using a linear path like IG~\cite{Sundararajan2017Axiomatic}, we get the attribution scores of $\mathcal{A}(\boldsymbol x^a, \boldsymbol 0) = \{3.5, 6.5\}$ and $\mathcal{A}(\boldsymbol x^b, \boldsymbol 0) = \{20, 19\}$, where $\boldsymbol 0 \in \mathbb R^2$ is a zero vector used by IG as the baseline. Clearly, these attribution scores are not only counter-intuitive, but also inconsistent.
\end{example}
\vspace{-1mm}

We verify that~\cite{Srinivas2019Full} rightly concludes that the  counter-intuitive behavior of IG (and other methods) for deep visual models occurs due to the violation of a property called \textit{weak dependence} - \textit{cf.}~Def.~(\ref{def:wd}). Simply put, an attribution method that depends weakly on model input, relies more strongly on the correct model weights to provide a credible attribution.  



\vspace{-1mm}
\begin{definition}[Weak dependence]\label{def:wd}
Consider a piece-wise linear model $F(.)$ encoded by `$p$'  pieces, defined over the same number of open connected sets $\mathcal{U}_i$ for $i\in[1,p]$ s.t. 
\begin{equation} \label{eq:wd2}
F(\boldsymbol x)=\left\{
\begin{array}{rcc}
\boldsymbol{w}_1^T \boldsymbol{x} + b_1, &      & \boldsymbol{x} \in \mathcal{U}_1\\
...\\
\boldsymbol{w}_p^T \boldsymbol{x} + b_p, &      & \boldsymbol{x} \in \mathcal{U}_p.
\end{array} \right.
\vspace{-0.7mm}
\end{equation}
For $F(\boldsymbol x)$, an attribution method  weakly depends on $\boldsymbol x$ when this dependence is only via the neighborhood set $\mathcal U_i$ of $\boldsymbol x$. 
\vspace{-1mm}
\end{definition}
The following important  proposition is also made in \cite{Srinivas2019Full}.
\vspace{-2mm} 

\begin{proposition}
``For any piece-wise linear function, it is impossible to obtain a saliency map\footnote{Termed `attribution map' in this work.} that satisfies both completeness and weak dependence on inputs, in general''~\cite{Srinivas2019Full}.
\end{proposition}

\vspace{-2mm}
\noindent{\underline{\bf P2: The baseline enigma:}} 
The baseline in the path attribution framework plays a key role in estimating the desired  scores. 
However, since path functions are not  originally  rooted in the vision literature~\cite{friedman2004paths}, a concrete definition of the baseline still eludes the path methods in the vision domain. Gradients of a model's  output w.r.t.~input are considered a useful  attribution measure in deep learning~\cite{Sundararajan2017Axiomatic}, \cite{Simonyan2014Deep}, \cite{baehrens2010explain}. 
However, it is also known that they can locally saturate for the important input  features~\cite{Sturmfels2020Visualizing}, \cite{shrikumar2017learning} because the prediction function of the model tends to flatten for those features. This compromises the credibility of the computed attributions, especially for the important features. The path attribution framework circumvents this problem with a non-local solution that integrates the gradients over a path from the baseline to the input - \textit{cf.}~Def.~(\ref{def:PbM}).
Here, the key assumption is that the baseline encodes the `absence' of the input feature. Only by satisfying this assumption, the solution of the path attribution framework is truly non-local.    

To emulate the feature absence, different path attribution methods for visual models employ different baselines. For instance, IG~\cite{Sundararajan2017Axiomatic} proposes  a black image as the baseline, whereas \cite{pan2021explaining} uses adversarial examples~\cite{akhtar2021advances}. A study in \cite{Sturmfels2020Visualizing} clearly shows that inappropriate encoding of feature absence in the baseline image can lead to severe  unintended effects on the  computed attribution scores.

\vspace{0.5mm}
\noindent{\underline{\bf P3: Ambiguous path features:}} Still largely unexplored in the literature is the intrinsic ambiguity of the features residing on the path specified by the path function - \textit{cf.}~Def.~(\ref{def:pf}). 
To preserve the properties of the framework, the path features must reside within the extremities of absence and presence of the input feature.  
However, partially owing to the problem \textbf{P2}, it remains  unknown if the paths defined by the existing methods in the vision domain are actually composed of the features that follow this intuition - see Fig.~\ref{fig:teaser}.    

In above, \textbf{P1} and \textbf{P2} are known but still  open problems, and \textbf{P3} is largely unexplored. Kapishnikov et al.~\cite{kapishnikov2021guided} came the closest to exploring \textbf{P3}, however they eventually altered the path  instead of addressing the features on the path. Besides the above issues, it is also known that gradient integration for path attribution can suffer from noise due to shattered gradients~\cite{balduzzi2017shattered}. However, this is known to be addressed well by computing Expected attribution scores using multiple baselines~\cite{Erion2021Improving}, \cite{Hooker2019Benchmark}, \cite{pan2021explaining}, \cite{Sturmfels2020Visualizing}.

\vspace{-2mm}
\section{Proposed fixes to the problems}
\label{sec:Fixes}
\vspace{-2mm}
Here, we  propose systematic fixes to the problems highlighted in \S~\ref{sec:Prob}. They are later combined to form a reliable path-based attribution scheme in \S~\ref{sec:method}.

\vspace{0.7mm}
\noindent{\underline{\bf F1: Avoiding counter-intuitive scores:}}
Problem \textbf{P1}  directly challenges the soundness of  path attribution framework for deep learning. Hence, we address that first. Building on Ex.~(\ref{example1}), we  provide Ex.~(\ref{example2}) that highlights the key intuition behind our proposed resolution of~\textbf{P1}.

\vspace{-1.5mm}
\begin{example}[Correct attributions scores]
\label{example2}
Consider the same $F(\boldsymbol x)$, $\boldsymbol x^a$ and $\boldsymbol x^b$ defined in Example~(\ref{example1}).
Let us choose a point $\boldsymbol x' = [3,3]$, s.t., $\boldsymbol x'\in \mathcal U_2$. When we use $\boldsymbol x'$ as the baseline instead of $\boldsymbol 0 \in \mathbb R^2$ and integrate using a linear path function,  we get $\mathcal A(\boldsymbol x^a, \boldsymbol x') = \{-6, -1.5\}$ and $\mathcal A(\boldsymbol x^b, \boldsymbol x') = \{4,1\}$. In general, we always get $abs(\frac{\mathcal A_1}{\mathcal A_2}) = 4$ whenever $\boldsymbol x' \in \mathcal U_2$,  which conforms to  the weights of the active piece of $F(\boldsymbol x)$ for $\boldsymbol x^a$ and $\boldsymbol x^b$. 
\end{example}
\vspace{-2mm}

In Example~(\ref{example2}), the key idea is to restrict the baseline to the same open connected set $\mathcal U_i$ to which the input belongs. We find that the path attribution framework always satisfies the weak dependence property along with completeness under this restriction. We make a formal proposition about it below. \textcolor{black}{Mathematical proof of the proposition is also provided in the  supplementary material of the paper.}

\vspace{-2mm}
\begin{proposition}\label{prop:2}
For a piece-wise linear function $F$,  path attribution satisfies both completeness and weak dependence simultaneously when the baseline $\boldsymbol x'$ and the input $\boldsymbol x$ belong to the same open connected set $\mathcal U_i$. 
\end{proposition}
\vspace{-2mm}






\vspace{0.7mm}
\noindent{\underline{\bf F2: Well-defined baseline:}}
To address the baseline ambiguity, we develop a concrete computational definition for this notion in Def.~(\ref{def:BL}). Our definition conforms to the original idea that the baseline  signifies feature absence~\cite{Sundararajan2017Axiomatic}. Additionally, we build on  Prop.~(\ref{prop:2}) to constrain the baseline to the open connected set of the neighbourhood of $\boldsymbol x$, which  helps in avoiding the counter-intuitive attributions.  
\vspace{-0.7mm}
\begin{definition}[Desired baseline]\label{def:BL}
Given a model $F$ and input $\boldsymbol x \in \mathbb R^n$, a desired baseline $\boldsymbol x' \in \mathbb R^n$  satisfies $||F(\boldsymbol x) - F(\boldsymbol x')||_2 \approx 0$, where~$\forall_{i \in \{1,...,n\}}$ $|x_i - x_{i}'| \geq \delta$. 
\end{definition}
\vspace{-.5mm}
In Def.~(\ref{def:BL}), the constraint $||F(\boldsymbol x) - F(\boldsymbol x')||_2 \approx 0$ encourages $\boldsymbol x'$ to use similar model weights as used by $\boldsymbol x$. A deep visual classifiers can be expressed as $F(x) = \mathcal C(\boldsymbol w_c, \mathcal R (\boldsymbol w_r, \boldsymbol x)): \boldsymbol x \rightarrow \boldsymbol l_x \in \mathbb R^L$, where $\mathcal C(.,.)$ and $\mathcal R(.,.)$ are respectively the  classification and   representation stages of the model. The constraint essentially imposes that the logit scores  for the baseline and the input are similar. For $\mathcal R(., .): \boldsymbol x \rightarrow \mathbb R^R$ , normally,  $R = \zeta L$  and $\zeta \approx 1$.  These conditions naturally promote  $\boldsymbol x'$ to use a similar weight set in $\mathcal{C}(.,.)$ as $\boldsymbol x$. To keep the flow of our discussion, we provide further  discussion on this phenomenon in the supplementary material. 

The external constraint $|x_i - x_{i}'|  \geq \delta$ imposes a minimum difference restriction over the baseline. We use it to enforce a computational analogue of  feature absence (explained further in \textbf{F3}) in $\boldsymbol x'$. When IG~\cite{Sundararajan2017Axiomatic} uses a black image as the baseline, the computed attribution map  assigns a zero score to the black features in the input. In fact, we observe that in general, whenever $x_i - x_{i}' \rightarrow 0$ for any feature, $\mathcal A_i(\boldsymbol x, \boldsymbol x_{i}') \rightarrow 0$ for the attribution method that uses a  linear path function. This is easily verifiable for IG by considering the term $(x_i - x_{i}')$ in Eq.~(\ref{eq:IG}). Our imposed constraint precludes this singularity. 

\begin{figure}[t]
    \centering
    \includegraphics[width = 0.35\textwidth]{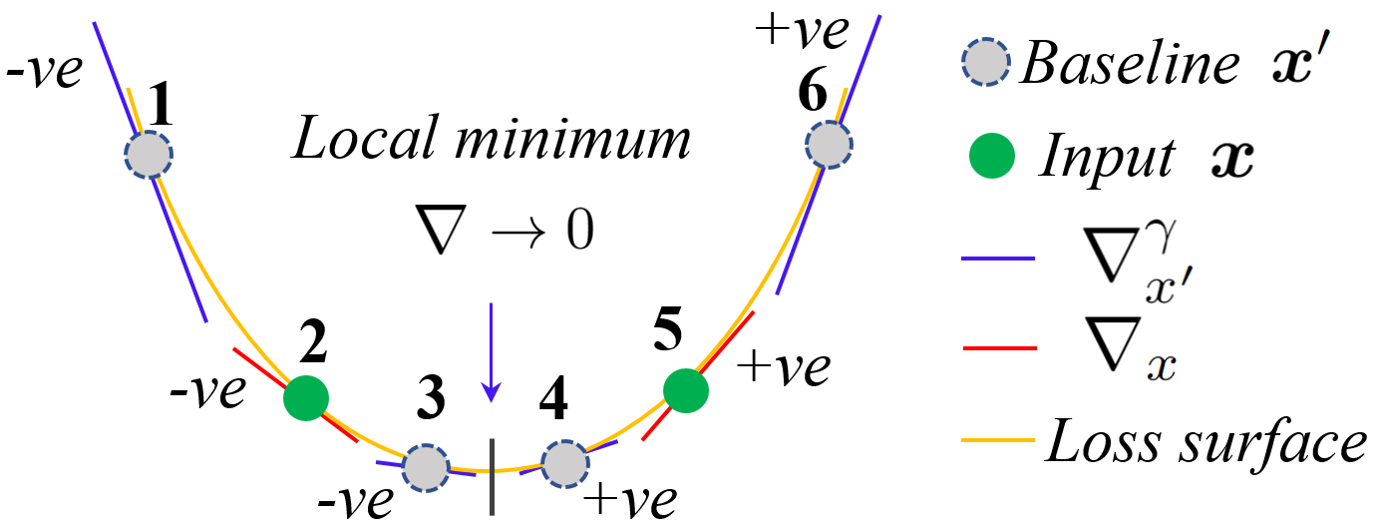}
    \caption{Input $\boldsymbol x$ is projected close to a local minimum of a well-trained model. By comparing the directions and magnitude of the gradients $\nabla_x$ and $\nabla_{x'}^{\gamma}$ of $\boldsymbol x$ and  $\boldsymbol x'$, we can estimate if $\boldsymbol x'$ (on path $\gamma$) encodes feature absence w.r.t.~$\boldsymbol x$ for the model in a computational sense. It occurs when $\text{sgn}(\nabla_x)\cdot \text{sgn}(\nabla_{x'}^{\gamma}) = 1$ and  $\text{abs}(\nabla_x) > \text{abs}(\nabla_{x'}^{\gamma})$. See text in \textbf{F3} for explanation.}
    \label{fig:grads}
    \vspace{-2.5mm}
\end{figure}

\vspace{1mm}
\noindent{\underline{\bf F3: Valid path features:}} To explain the valid path features, we first need to further explain our computational view of the feature absence. 
For that, refer to Fig.~\ref{fig:grads}, which plots a hypothetical smooth loss surface assuming a well-trained model. As  the model is well-trained, the (projection of) input $\boldsymbol x$ (say at location 2) is close to a local minimum. With respect to the model, a higher (computational) `presence' of the feature  in the baseline  asserts that its location  is even closer to the local minimum than $\boldsymbol x$, e.g., at location~3. Conversely, a higher feature `absence' will require $\boldsymbol x'$ to reside farther from $\boldsymbol x$, e.g., at location 1. For a smooth surface, gradients flatten near the optima and remain relatively steep elsewhere. 
Hence, we observe that \textit{by comparing the magnitudes of the gradients for $\boldsymbol x$ and $\boldsymbol x'$, we can identify if  $\boldsymbol x'$ encodes feature absence, especially when $||F(\boldsymbol x) - F(\boldsymbol x')||_2$ is small}. We denote the gradient for $\boldsymbol x$ by $\nabla_x$, and for $\boldsymbol x'$ by $\nabla^{\gamma}_{x'}$ in the figure, where $\gamma$ restricts $\boldsymbol x'$ to be on the path defined by the path function $\gamma(\alpha)$\footnote{We eventually choose a linear path for our method. In that case, $\nabla^{\gamma}_{x'} = (x_i -x'_{i})\cdot \nabla_{x'}$, where $\nabla_{x'}$ is the model gradient w.r.t.~$\boldsymbol x'$. Also notice that to keep the discussion flow, we treat  $\boldsymbol x'$ in Fig.~\ref{fig:grads}, and the related text to be `any' point on the path between the baseline and the input - not just as the baseline. This changes in the formal definition in Def.~(\ref{def:vpf}).}.  

Looking closely, the above observation fails when $\boldsymbol x'$ picks location 4 instead of 3, which still has a smaller gradient magnitude  than location 2, however it may not represent a larger feature presence. This is because, for our observation  to hold, the local minimum must be approached from (not towards) the input $\boldsymbol x$. Luckily, we can identify that 4 is not on the same side as $\boldsymbol x$ by comparing the sign of the gradient, i.e.,~$\text{sgn}(\nabla_{x'}^{\gamma})$, at that location with $\text{sgn}(\nabla_x)$ at location 2. From Fig.~\ref{fig:grads}, it is clear  that the observation in the preceding paragraph holds in general when we  additionally impose $\text{sgn}(\nabla_x)\cdot \text{sgn}(\nabla_{x'}^{\gamma}) = 1$.

The example in Fig.~\ref{fig:grads} may  at first seem contrived.  However, it is fully generalizes to any $F$ for which $\frac{\partial F(\gamma(\alpha))}{\partial x_i}$ exists everywhere - \textit{cf.}~Def.~(\ref{def:pf}). Hence, we can identify the valid features on the path defined by our path function $\gamma$  as
\vspace{-0.7mm}
\begin{definition}[Valid path features]\label{def:vpf} A feature $\tilde{x}_i \in \mathbb R$ on the path $\gamma(\alpha)$ - cf.~Def.~(\ref{def:pf}) - defined by the  input $x_i$ and a baseline $x'_i$ is a valid path feature when $\text{sgn}(\nabla_{x_i})\cdot \text{sgn}(\nabla^{\gamma}_{\tilde {x}_i}) = 1$ and $\text{abs}(\nabla_{x_i})> \text{abs}(\nabla^{\gamma}_{\tilde {x}_i}).$
\end{definition}


\vspace{-3mm}
\section{Proposed attribution method}
\label{sec:method}
\vspace{-1mm}
With Def.~(\ref{def:BL}), we specified a baseline that precludes the  counter-intuitive attributions under Prop.~(\ref{prop:2}) while also conforming to a sensible computational analogue of feature absence for the visual models. Def.~(\ref{def:vpf}) provides a verification check to ensure that the features on our path indeed reside between the extremities of the input feature absence and its   presence\footnote{Computationally, a path-constrained feature that overshoots its presence in the input  becomes an invalid feature as per Def.~(\ref{def:vpf}).}.  We now describe our procedure to combine these insights into a reliable path-based attribution method.

\vspace{0.5mm}
\noindent{\underline{\bf Baseline computation:}} The desired baseline in Def.~(\ref{def:BL}) leads to the following optimization problem. 
\vspace{-0.7mm}
\begin{align}
 \underset{x'}{\min}|| F_{\text{logits}}(\boldsymbol x) - F_{\text{logits}}(\boldsymbol x')||_2~ \text{s.t.}~ \underset{i}{\min}~|x_i- x'_i| \geq \delta,   
 \label{eq:BLE}
 \vspace{-0.7mm}
\end{align}
where `logits' indicates the use of the model logit scores. To solve this, we propose Algo.~(\ref{alg:BL}). The  idea of Algo.~(\ref{alg:BL}) is to first create an initial estimate $\boldsymbol x^p$ of $\boldsymbol x'$ under the transformations $\psi(.)$, such that $\boldsymbol x^p$ differs from $\boldsymbol x$  in both input and output spaces. We use a  Gaussian blur for that  purpose.  
Then, in lines \ref{1:l2} - \ref{1:l7} of Algo.~(\ref{alg:BL}), we gradually alter $\boldsymbol x^p$ to bring it close to $\boldsymbol x$ in the model output space, while maintaining the constrain ${\min_i}~|x_i- x^p_i| \geq \delta$ in the input space. Here, logit scores are used as the output map of $F$.  
On line~\ref{1:l3}, alteration to $\boldsymbol x^p$ is guided by Lemma~(\ref{lemma1}), which provides us with a desirable direction of altering  $\boldsymbol x^p$ that can efficiently achieve our objective. We take small steps in that  direction with a step size $\eta$. On lines \ref{1:l4} - \ref{1:l6}, we ensure that $\boldsymbol x^p$ abides by $|x_i- x^p_i| \geq \delta$ after each alteration. Line \ref{1:l7} brings the image back to the valid dynamic image range for the model $F$ by clipping it.


\vspace{-1.5mm}
\begin{lemma}\label{lemma1}
For $F(.)$ with cross-entropy loss, $F(\boldsymbol x^p)$ can approach  $F(\boldsymbol x)$ by stepping in the direction $-\text{sgn}(\nabla_{x^p}F)$.  
\end{lemma}
\vspace{-1.5mm}
\noindent{\bf Proof:} \textit{For  $F(.)$ with a cross-entropy loss $\mathcal J(.,.)$, $F(\boldsymbol x^p) \rightarrow F(\boldsymbol x)$ requires maximizing $\log \left( p (F(\boldsymbol x)| \boldsymbol x^p \right)$, which is the same as  stepping in the direction sgn$\left(\nabla_{x^p} \log ( p (F(\boldsymbol x)| \boldsymbol x^p )\right)$. This direction is the same as $\text{sgn}\left(- \nabla_{x^p} \mathcal J(F(x), \boldsymbol x^p) \right)$ or $-\text{sgn}(\nabla_{x^p}F)$ following our short-hand notation.}

\begin{algorithm}[t]
 \caption{ComputeBaseline}
 \label{alg:BL} 
 \begin{algorithmic}[1]
 \renewcommand{\algorithmicrequire}{\textbf{Input:}}
 \renewcommand{\algorithmicensure}{\textbf{Output:}}
 \REQUIRE  Image $\boldsymbol x \in \mathbb R^n$, model $F$, Blur kernel size $\sigma$, Gradient step size $\eta$, Thresholds $\epsilon$, $\delta$.
 \ENSURE Baseline $\boldsymbol x' \in \mathbb R^n$.
 \STATE \label{1:l1}$\boldsymbol{x}^{p} \leftarrow \psi(\boldsymbol x, \sigma)$ ~~~~~~~\textcolor{gray}{//assert $\boldsymbol{x}^{p} \neq \boldsymbol x$}
\WHILE{$||F_{\text{logits}}(\boldsymbol x) - F_{\text{logits}}(\boldsymbol x^p)||_2 > \epsilon$} \label{1:l2} 
\STATE  \label{1:l3} $\boldsymbol x^p \leftarrow \boldsymbol x^p$ - $\eta$ $\cdot$ sgn $\left( \nabla_{x^p}F \right)$ 
\FOR{$i = 1~~\text{to}~~n$}  \label{1:l4}
\IF{$ |x_i - x^p_i| < \delta$}   \label{1:l5}
\STATE  \label{1:l6} $x^p_i \leftarrow -\text{sgn}( x_i - x^p_i) \cdot \delta + x^p_i $
\ENDIF
\ENDFOR
\STATE  \label{1:l7} $\boldsymbol x^p \leftarrow \text{Clip} (\boldsymbol x^p)$
\ENDWHILE
\STATE  \label{1:l8} $\boldsymbol x' \leftarrow \boldsymbol x^p$
 \end{algorithmic}
 \end{algorithm}
 
 \vspace{1mm}
 \noindent{\underline{\bf Gradients integration:}}
 Algorithm~(\ref{alg:Main}) summarizes our  technique to compute the attribution map with gradient integration. For clarity, the text below describes it in a non-sequential manner. On line \ref{2:l3} of Algo.~(\ref{alg:Main}), we obtain the desired baseline image by calling Algo.~(\ref{alg:BL}). Using this  baseline, line~\ref{2:l6} computes the features that reside on the path between the baseline and the input, employing a step size sampled from a uniform distribution. We also use a linear path function similar to IG~\cite{Sundararajan2017Axiomatic} - \textit{cf.}~\S~\ref{sec:Prob}. This allows our method to inherit the axiomatic properties of the canonical path method. We discuss this aspect in more detail in the supplementary material of the paper while describing the theoretical properties of the path attribution framework. 
 
 After computing the path features, their gradients on the path are estimated on line \ref{2:l7}, and the checks specified by Def.~(\ref{def:vpf}) are performed on line \ref{2:l9}.  It is straightforward to show that under the Riemman approximation of the integral, gradients for a feature $x_i$ on a linear path can be integrated as $(x_i - x'_i) \times \frac{1}{m} \sum_{i = 1}^m \nabla_{\tilde x_i} F$, where $ \nabla_{\tilde x_i} F$ is the model gradient w.r.t.~the valid path feature $\tilde{x}_i$. On line~\ref{2:l10}, accumulation of the gradients of the valid features, i.e.,~$\sum_i \nabla_{\tilde x_i} F$, is performed, which is used for the Reimann approximation of integration on line~\ref{2:l11} of the algorithm.
 
 Besides the above, an outer \textit{for-loop} can be observed in Algo.~(\ref{alg:Main}). This loop allows us to use multiple baselines in our path-based attribution. Using multiple baselines can be beneficial in suppressing the noise in accumulated gradients~\cite{Erion2021Improving}.  Erion et al.~\cite{Erion2021Improving} used a Monte Carlo approximation of the integral in their technique to leverage multiple baselines. Inspired, we also use the same approximation, which requires Algo.~(\ref{alg:Main}) to estimate the eventual integral as $\underset{\boldsymbol x' \sim \mathcal B}{\mathbb E} [(\boldsymbol x - \boldsymbol x') \nabla_{\tilde{x}}F]$, where $\mathcal{B}$ is the distribution over the proposed baseline. Mathematically, the noted Expectation value leads us to averaging over the gradients which are integrated in the inner \textit{for-loop}. This is accomplished with the outer loop. It is noteworthy that we allow multiple baselines in our method mainly to suppress any potential noise due to the shattered gradients problem. Otherwise, the inner \textit{for-loop}, along the baseline computation in Algo.~(\ref{alg:BL}), already accounts for the fixes \textbf{F1} - \textbf{F3} discussed in \S~\ref{sec:Fixes}.  
 
 The hyper-parameters of the proposed method are handles over intuitive concepts, which makes selecting their values straightforward. We provide a detailed discussion about them in the supplementary material of the paper, where we also demonstrate that the computed attribution scores are largely insensitive to a wide range of sensible values of these parameters.


 
\newcommand{\Exp}[1]{\underset{#1}{\mathbb E}}
\begin{algorithm}[t]
 \caption{Path integration}
 \label{alg:Main} 
 \begin{algorithmic}[1]
 \renewcommand{\algorithmicrequire}{\textbf{Input:}}
 \renewcommand{\algorithmicensure}{\textbf{Output:}}
 \REQUIRE  Image $\boldsymbol x \in \mathbb R^n$, model $F$, $\#$ of baselines $B$, Total steps $K$, param = $\{\eta, \epsilon, \delta \}$, Blur kernel sizes $\{\sigma_b\}_{b = 1}^{B}$
 \ENSURE Attribution map $\mathcal A$
 \STATE \label{2:l1}Initialize:  $\text{gradAcc} = \boldsymbol 0 \in \mathbb R^n$, $\boldsymbol\rho \leftarrow \nabla_x F$ 
 \FOR {$b$ =  $1$ to $B$} \label{2:l2}
 \STATE \label{2:l3} $\boldsymbol x' \leftarrow $ ComputeBaseline$(\boldsymbol x, F, \sigma_b,  \text{param})$
 \STATE \label{2:l4} count  $\leftarrow \boldsymbol 0 \in \mathbb R^n$,  $\boldsymbol \varrho \leftarrow \boldsymbol 0 \in \mathbb R^n$
 \FOR {$k$ = $1$ to $\lfloor \frac{K}{B} \label{2:l5} \rfloor$}
 \STATE \label{2:l6} $ \tilde{\boldsymbol x} \leftarrow \boldsymbol x' + \alpha_k (\boldsymbol x - \boldsymbol x')$ s.t.~$\alpha_k \sim \text{uniform}(0,1)$
 \STATE \label{2:l7} $\tilde{\boldsymbol\rho} \leftarrow (\boldsymbol x - \tilde{\boldsymbol x})\cdot~\nabla_{\tilde{x}} F$
 \FOR{ $i$ = $1$ to $n$} \label{2:l8}
 \IF{sgn$(\boldsymbol \rho_i)$ = sgn$(\tilde{\boldsymbol \rho}_i)$ $\wedge$ $|\boldsymbol \rho_i| > |\tilde{\boldsymbol \rho}_i|$} \label{2:l9}
 \STATE \label{2:l10} $\boldsymbol \varrho_i \leftarrow \boldsymbol \varrho _i+ \nabla_{\tilde{x}_i} F$, 
 count$_i \leftarrow $ count$_i + 1$
 \ENDIF
 \ENDFOR
 \ENDFOR
  \STATE \label{2:l11} $\text{gradAcc} \leftarrow \text{gradAcc} + (\boldsymbol x - \boldsymbol x') \boldsymbol{\varrho}\cdot/$count 
 \ENDFOR
\STATE \label{2:l12} $\mathcal A \leftarrow$ gradAcc $/ B$
 \end{algorithmic}
 \end{algorithm}
 
 \vspace{-2mm}
\section{Experiments}
\label{sec:EE}
\vspace{-1mm}
This paper contributes to the path-based model interpretation paradigm, hence its experiments are specifically designed to show improvements to the path attribution framework with the newly provided insights. Indeed, there are also other post-hoc interpretation techniques besides path methods, \textit{cf.}~\S~\ref{sec:RW}. However, they are not axiomatic, which renders their comparison with the path methods injudicious. This is particularly true for quantitative comparisons because to-date there is no mutually agreed upon metric that is known to comprehensively quantify the correctness of  attribution maps.  It is emphasized that the intent of our evaluation in not to claim new state-of-the-art on performance metrics, which are disputed in the first place. Rather, we use empirical results as an evidence that our theoretical insights positively contribute to the path attribution framework.      

\vspace{0.75mm}
\noindent {\bf Insertion/Deletion evaluation on ImageNet:} Among the most commonly used quantitative evaluation metrics for the post-hoc interpretation methods, are the insertion and deletion game  scores~\cite{petsiuk2018rise}. In our evaluation, the insertion game inserts the most important pixel (as computed by the method) first and records the change in the model output. The deletion game conversely records the score by removing the most important pixel first. We conduct insertions and deletions for all the pixels and compute the Area Under the Curve (AUC) of the output change with the pixel insertion/deletion.
For insertion, a larger AUC is more desirable, which is opposite for the deletion. It is easy to see that the two metrics do not capture the full picture of method performance individually. Hence, we combine them by reporting the AUC of the difference between the insertion and deletion scores in our results, where the larger differences become more desirable. This is a more comprehensive metric. 

In Table~\ref{tab:ID_ImgNet}, we summarize the results on three popular ImageNet models. As the baseline method, we chose the canonical path attribution technique, i.e.,~Integrated Gradients (IG)~\cite{Sundararajan2017Axiomatic}. We also implement IG, using different path baselines. IG (G) uses Gaussian noise instead of black image as the path baseline, whereas IG (A) uses the average pixel value of the input as the baseline. For each model, the results are averaged over $2,500$ images from the ImageNet validation set. We also include the existing relevant methods Expected Gradient (EG)~\cite{Erion2021Improving} and Adversarial Gradient Integration (AGI)~\cite{pan2021explaining} for benchmarking.  
\begin{table}[t]
\setlength{\tabcolsep}{0.4em}
    \centering
    \begin{tabular}{l|c|c|c} \hline
        \cellcolor{green!25} Method &  \cellcolor{green!25} ResNet-50 & \cellcolor{green!25} DenseNet-121  & \cellcolor{green!25} VGG-16 \\ \hline
         IG~\cite{Sundararajan2017Axiomatic}         &  0.3711              & 0.5042                & 0.2532   \\ \hline
         IG (G)      &  0.2997 \reduction{-19.2}              & 0.4128 \reduction{-18.1}             & 0.1609  \reduction{-36.4}  \\ \hline
         IG (A)      &  0.4653 \improvement{+25.4}             & 0.5493 \improvement{+8.9}              & 0.3387  \improvement{+33.7}  \\ \hline
         AGI~\cite{pan2021explaining}          &  0.3995 \improvement{+7.6}              & 0.4549  \reduction{-9.7}              & 0.2419   \reduction{-4.4} \\ \hline 
         EG~\cite{Erion2021Improving}          &  0.4004 \improvement{+7.90}              & 0.5171  \improvement{+2.5}              & 0.2727   \improvement{+7.69} \\ \hline 
         Our         &  \textbf{0.5311} \improvement{+43.1}     & \textbf{0.6228} \improvement{+23.5} & \textbf{0.4023} \improvement{+58.8}   \\ \hline
    \end{tabular}
\caption{ImageNet~\cite{deng2009imagenet} AUC difference between Insertion-Deletion scores~\cite{petsiuk2018rise}. Percentage gain over Integrated Gradient (IG)~\cite{Sundararajan2017Axiomatic} is also given. IG (G) and IG (A) respectively use Gaussian noise and Average pixel value of the input as the baseline.}
    \label{tab:ID_ImgNet}
    \vspace{-1mm}
\end{table}
\begin{table}[t]
    \centering
    \setlength{\tabcolsep}{0.4em}
 \begin{tabular}{l|c|c|c} \hline
        \cellcolor{green!25} Method &  \cellcolor{green!25} RN-50 (0.77) & \cellcolor{green!25} RN-34 (0.75)  & \cellcolor{green!25} RN-18 (0.64) \\ \hline
         IG~\cite{Sundararajan2017Axiomatic}         &  0.3711              & 0.3854                & 0.2481   \\ \hline
         Our       &  \textbf{0.5311}\improvement{+43.1}              & \textbf{0.5185}\improvement{+34.5}                & \textbf{0.3675}\improvement{+48.1}   \\ \hline
    \end{tabular}
    \caption{Performance gain over IG for different ResNet (RN) variants. Average confidence score of models are noted in parenthesis.}
    \label{tab:RN}
    \vspace{-3mm}
\end{table}

In Table~\ref{tab:ID_ImgNet}, we ensure that all the methods use the same images and models, and also take the same number of steps from the baseline image to the input. This allows a fully transparent comparison. For all the methods, we allow $150$ steps. Since our technique enables the use of multiple baselines, we use $3$. The same number of baselines and steps are used for EG~\cite{Erion2021Improving} and AGI~\cite{pan2021explaining}. The reported results also include the percentage gains of each technique over IG. Since IG is the canonical path attribution method~\cite{Sundararajan2017Axiomatic}, it provides the perfect baseline to establish any positive development for the path attribution framework. It can be noticed that our method achieves remarkable gains, with up to $58.8\%$ improvement for VGG-16.  

There are a few reportable interesting observations related to the results in Table~\ref{tab:ID_ImgNet}. We noticed that the average confidence scores of ResNet-50, DenseNet-121 and VGG-16 in our experiments were $0.77$, $0.81$ and $0.69$, respectively. The underlying pattern  is exactly the opposite to that of the gains we achieved over IG with our method in Table~\ref{tab:ID_ImgNet}. Indicating that IG may have a tendency to perform sub-optimally (relatively speaking) for the less confident models - as adjudged by the insertion/deletion game scores. To further verify that, we report the results of an additional experiment with ResNet (RN) variants in Table~\ref{tab:RN}. Whereas IG gained some grounds for ResNet-34, it again performed relatively poorly for ResNet-18, which shows its general tendency to get affected by the model confidence strongly. Our method consistently performed excellent for all the ResNet variants.

Another interesting observation we made related to the results in Table~\ref{tab:ID_ImgNet}, was about the performance of EG~\cite{Erion2021Improving} and AGI~\cite{pan2021explaining}. Whereas we use author-provided codes for these methods, we match the hyper-parameters with IG and our method, and remove any other pre-/post-processing of the computed maps which is not used by IG. For instance, we remove thresholding of AGI, which does not conform to the axioms of path attribution. This places AGI on equal grounds with IG. As can be seen in Table~\ref{tab:ID_ImgNet}, AGI does not perform too well on equal grounds with IG. We use the best performing variant of AGI in our experiments, which was achieved with PGD attack~\cite{Madry2018Towards}. For EG, we use 3 baselines with 50 steps to fairly match it with our method. This variant performed almost similar to using 150 baselines with 1 step for each baseline.    

\vspace{0.75mm}
\noindent {\bf Insertion/Deletion evaluation on CIFAR-10:} As compared to $224\times224$ grid size of ImageNet samples, CIFAR-10~\cite{CIFAR10} has $32\times 32$ image grid size. Image size has direct implications for the quantitative metrics of insertion and deletion games. Hence, in Table~\ref{tab:CIFAR10}, we also report performance of our method on 1000 images of CIFAR-10 validation set. In the table, we only include the top performing approaches from Table~\ref{tab:ID_ImgNet}. On  CIFAR-10 images, IG already performed considerably well under insertion/deletion score metrics. Nevertheless, our method still provided a considerable relative gain over IG consistently. It is noteworthy that our observation regarding the relation between the relative gain of our method over IG and the model confidence scores also holds in the CIFAR-10 experiments. 

\begin{table}[t]
    \centering
    \begin{tabular}{l|c|c|c} \hline
        \cellcolor{green!25} Method &  \cellcolor{green!25} ResNet-50 & \cellcolor{green!25} DenseNet-121  & \cellcolor{green!25} VGG-16 \\ \hline
         IG~\cite{Sundararajan2017Axiomatic} & 0.5502 & 0.4693 & 0.4873  \\ \hline
         IG (A) & 0.5619\improvement{+2.1} & 0.5039\improvement{+7.4} & 0.4964\improvement{+1.8} \\ \hline
         Our & \textbf{0.5889}\improvement{+7.1} & \textbf{0.5448}\improvement{+16.1} & \textbf{0.5120}\improvement{+5.0} \\ \hline
    \end{tabular}
    \caption{CIFAR-10~\cite{CIFAR10} AUC difference between Insertion-Deletion scores~\cite{petsiuk2018rise}. Average confidence scores of ResNet-50, DenseNet-121  and VGG-16 on the images are 0.81, 0.74 and 0.79, respectively. IG (A) uses average pixel value of input as  baseline.}
    \label{tab:CIFAR10}
    \vspace{-4mm}
\end{table}

\begin{SCfigure*}
  \centering
  \caption{Sensitivity-N~\cite{ancona2017towards} analysis on ImageNet models. Pearson Correlation Coefficient (PCC) between the sum of the attributions and output variations under different sampling set size ratios are plotted. Larger values are more desirable. A considerable gain is achieved by our method over IG~\cite{Sundararajan2017Axiomatic}.} 
  \includegraphics[width=0.74\textwidth]%
    {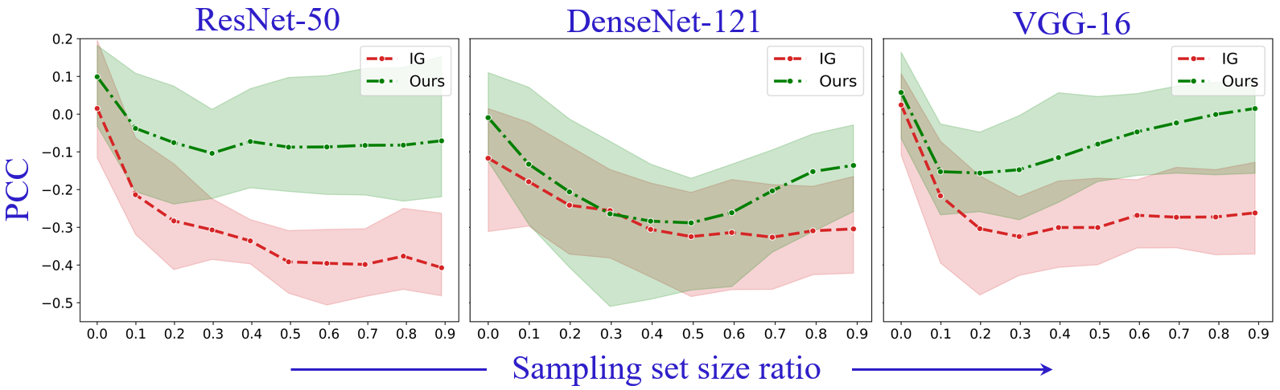}
    \label{fig:SensitivityN}
\end{SCfigure*}

\begin{figure*}[t]
    \centering
    \includegraphics[width =0.9\textwidth]{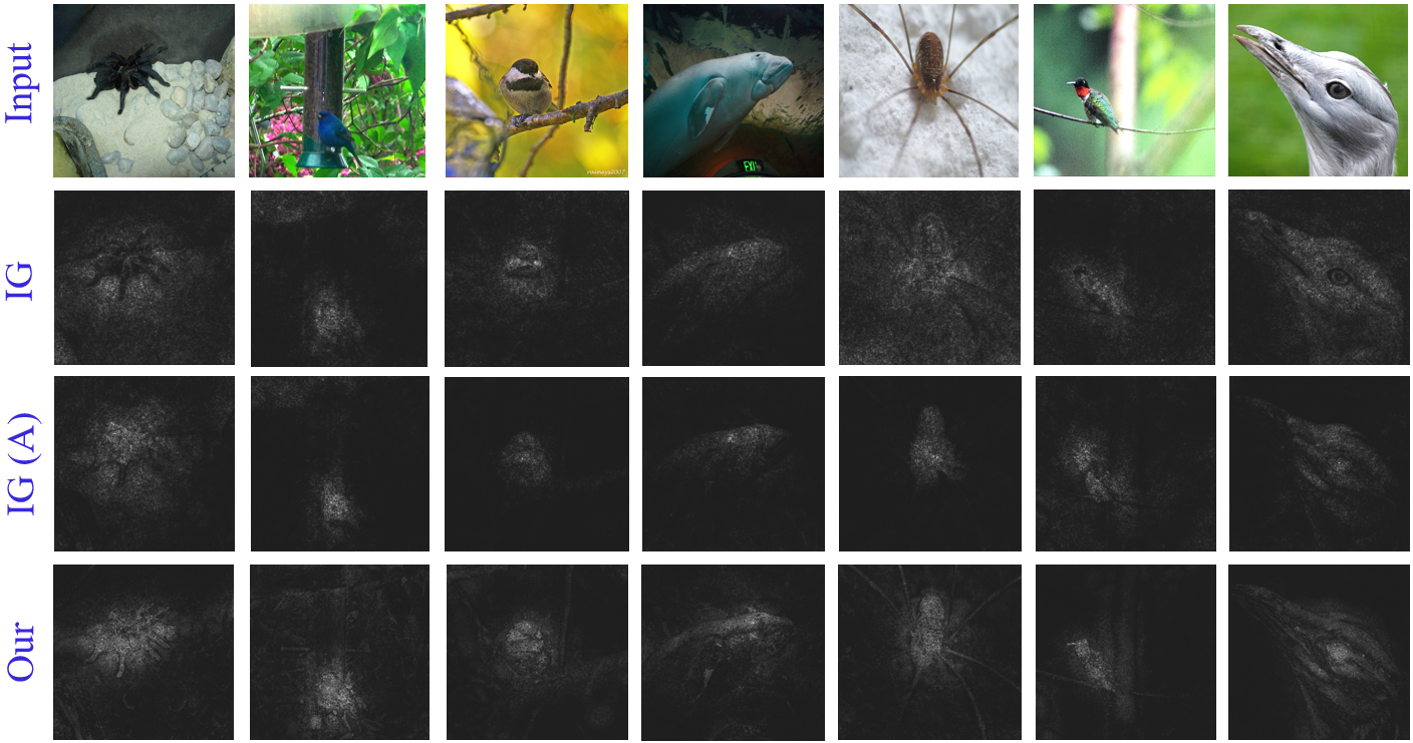}
    \caption{Representative visualizations of ImageNet image attributions with VGG-16 predictions. Best viewed enlarged on screen.}
    \label{fig:visual}
    \vspace{-3mm}
\end{figure*}

\vspace{0.7mm}
\noindent{\bf Sensitivity-N evaluation on ImageNet:} Though it is common to evaluate performance of attribution methods under a single quantitative metric~\cite{kapishnikov2021guided}, \cite{pan2021explaining}, we further evaluate our approach with Sensitivity-N~\cite{ancona2017towards} to thoroughly establish its contribution to the path attribution framework. The \textit{computationally intensive} Sensitivity-N metric comprehensively verifies that the model output is sensitive to the pixels considered important by the attribution method.  For any feature subset of $\boldsymbol x$, i.e., $\boldsymbol x_{\text{sub}} = [ x_1,  x_2,..., x_k] \subseteq \boldsymbol x$, this metric requires that the relation $\sum_i^k \mathcal A_i = f(\boldsymbol x) - f(\boldsymbol x_{[x_{\text{sub}}=0]})$ holds. Whereas no method is expected to fully satisfy this condition due to practical reasons, the metric is still useful. To put it into the practice, we vary the feature fraction in $\boldsymbol x_{\text{sub}}$ in the range [0.01, 0.9], and compute the Pearson Correlation Coefficient between $\sum_i^m \mathcal A_i$ values and the output variations. A larger correlation indicates better performance. 

We plot the results of Sensitivity-N for the  ImageNet model interpretations in Fig.~\ref{fig:SensitivityN}. 
It is observable that as compared to the canonical method IG, our method generally performs considerably better under this metric as well. We emphasize that Sensitivity-N is a computationally demanding  comprehensive metric. Our results on ImageNet sized images for this metric conclusively establish the efficacy of the proposed method.  

\vspace{0.7mm}
\noindent{\bf Qualitative results:} In Fig.~\ref{fig:visual}, we show representative qualitative results for random samples using VGG-16 model for ImageNet. Results of only the top-performing methods in Table~\ref{tab:ID_ImgNet} are included due to space restrictions. Attribution scores are encoded as gray-scale variations, where brighter pixels represent larger attribution scores. We also provide more results in the supplementary material. It is clear from the results that our method does not face the problem of assigning lower scores to the dark pixels of the object. Our maps are also less noisy and indeed assign large attributions to the foreground object. We do not observe counter-intuitive behavior of attributions in our method.
These properties can be directly attributed to the design of our method which pays special attention to   incorporate them. 


\begin{table}
    \centering
    \setlength{\tabcolsep}{0.1em}
    \begin{tabular}{l|c|c|c} \hline
        \cellcolor{green!25} Method &  \cellcolor{green!25} ResNet-50 & \cellcolor{green!25} Dense-121  & \cellcolor{green!25} VGG-16 \\ \hline
         Our + IG\_Baseline & 0.3809& 0.5112&0.2552 \\ \hline
         Our (Single\_Baseline) &  0.5254              & 0.6192                & 0.3602   \\ \hline
            Our (Proposed)        &  \textbf{0.5311}      & \textbf{0.6228}  & \textbf{0.4023}    \\ \hline
    \end{tabular}
    \caption{Contribution of the proposed baseline and integration process to the final results on ImageNet models. AUC differences reported for insertion/deletion game scores.}
    \label{tab:ablation}
    \vspace{-5mm}
\end{table}

\noindent{\bf Further results:} The ablation analysis in Table~\ref{tab:ablation} shows that both the proposed baseline and the  path feature integration process contribute positively to the overall performance of our method. In the supplementary material, we provide further results demonstrating the effects of hyper-parameter settings on the performance. The key finding for those results is that we can even improve our performance further by allowing more steps on the path, and our results are generally insensitive to the hyper-parameters values  in reasonable ranges. Computational and memory requirements of our method also remain comparable to those of IG. We also provide details in that regard in the supplementary material.  

\vspace{-3mm}
\section{Conclusion}
\vspace{-1.5mm}
Using theoretical guidelines, this paper pinpointed the sources of three shortcomings of the path attribution framework that compromise its reliability as an interpretation tool for the  deep visual models. It  proposed fixes to these problems such that the framework becomes fully conformant to the original game-theoretic intuitions that govern its much desired axiomatic properties. We combined these fixes into a concrete path attribution method that can compute reliable explanations of deep visual models. The claims are also  established by an extensive empirical evidence to explain a range of deep visual classifiers. 

\vspace{3mm}
\noindent{\bf  Acknowledgements}\\
Dr.~Naveed Akhtar is recipient of an Office of National Intelligence National Intelligence Postdoctoral Grant (project number NIPG-2021-001)
funded by the Australian Government.

\balance
{\small
\bibliographystyle{ieee_fullname}
\bibliography{Submission.bib}
}

\end{document}